# Developing Successful Shared Tasks on Offensive Language Identification for Dravidian Languages

Bharathi Raja Chakravarthi[-], Dhivya Chinnappa[♦], Ruba Priyadharshini[*], Anand Kumar Madasamy[-ft], Sangeetha Sivanesan[ŧ], Subalalitha Chinnaudayar Navaneethakrishnan[♦], Sajeetha Thavareesan[tl], Dhanalakshmi Vadivel[$], Rahul Ponnusamy[O], Prasanna Kumar Kumaresan[O]

[-] Insight SFI Research Centre for Data Analytics, National University of Ireland Galway, Galway, Ireland
[♦] TR Labs, Thomson Reuters, USA
[*] ULTRA Arts and Science College, Madurai, Tamil Nadu, India
[-ft] National Institute of Technology Karnataka Surathkal, Karnataka, India
[tl] Eastern University, Sri Lanka
[ŧ] National Institute of Technology Tiruchirappalli, Tamil Nadu, India
[♦] SRM Institute of Science and Technology, Chennai, Tamil Nadu, India
[O] Indian Institute of Information Technology and Management-Kerala
bharathi.raja@insight-centre.org, dhivya.chinnappa@thomsonreuters.com, rubapriyadharshini.a@gmail.com, m_anandkumar@nitk.edu.in, sangeetha@nitt.edu, subalaln@srmist.edu.in, sajeethas@esn.ac.lk, dhanagiri@gmail.com, rahul.mi20@iiitmk.ac.in, prasanna.mi20@iiitmk.ac.in

## Abstract

With the fast growth of mobile computing and Web technologies, offensive language has become more prevalent on social networking platforms. Since offensive language identification in local languages is essential to moderate the social media content, in this paper we work with three Dravidian languages, namely Malayalam, Tamil, and Kannada, that are under-resourced. We present an evaluation task at FIRE 2020-HASOC-DravidianCodeMix and DravidianLangTech at EACL 2021, designed to provide a framework for comparing different approaches to this problem. This paper describes the data creation, defines the task, lists the participating systems, and discusses various methods.

## 1. Introduction

In the digital age, social media plays an important role in online communication, allowing users to create and share material while also giving accessible means to express their views and thoughts on anything at any time (Edosomwan et al. 2011). However, with the advent of social media, platforms such as YouTube, Facebook, and Twitter not only aided in information sharing and networking, but they also became a place where people were targeted, defamed, and marginalized based solely on their physical appearance, religion, or sexual orientation (Keipi et al. 2016; Benikova et al. 2018; Pamungkas et al. 2020). Because of the rising proliferation of dangerous and offensive information on social networking sites over the last decade, many experts have focused on the systematic identification of hate speech (Kumar et al. 2018) and offensive language identification (Zampieri et al. 2020; Mandl et al. 2020; Chakravarthi et al. 2021).

In linguistics, code-mixing is mixing between two or more languages in the same utterance. In a multilingual community, code-mixing is common, and code-mixed writings are occasionally produced in non-native scripts (Barman et al. 2014; Bali et al. 2014; Jose et al. 2020). People find it simpler to converse when two or more languages are mixed together or when their original tongue is written in Latin character (Chittaranjan et al. 2014; Chakravarthi et al. 2020d). Due to the growth of social media platforms across the world and the possibility of writing content without any moderation, users write content in multilingual code-switching without grammatical restrictions and using non-native scripts (Rudra et al. 2016; Chakravarthi et al. 2019). Due to historical reasons and present computer keyboard layouts, user-generated material is frequently typed in the Latin script in India, Sri Lanka and Singapore, like multilingual countries. As a result, the bulk of user-generated data for these Dravidian languages is code-mixed (Priyadharshini et al. 2020; Jose et al. 2020). This explosion of code-mixed user-generated content in social media platforms also makes it challenging to identify offensive content.

Tamil, Malayalam, and Kannada are Dravidian languages spoken by around 220 million people in the Indian subcontinent, Singapore, and Sri Lanka ( Chakravarthi et al. 2020c). Although considerable progress has been achieved in identifying offensive English language and hate speech, most research has mostly concentrated on identifying the abusive and offensive language in monolingual settings. This subject still appears to be at a very early stage of research for under-resourced languages such as Tamil, Malayalam, and Kannada, which lack tools and datasets (Chakravarthi et al. 2020a; Thavareesan and Mahesan 2020) as well as it is code-mixed social media text.

As a result, for the first time, a multilingual Dravidian corpus collected from similar themes has been generated and made available to the research community to identify offensive language. In fact, the first large-scale freely available resource for offensive language identification in Dravidian languages were the datasets that we developed for Hate Speech and Offensive Content (HASOC)-Offensive Language Identification-DravidianCodeMix-2020. This paper builds on top of our work Chakravarthi et al. (2020b), which was further extended for DravidianLangTech-2021 Chakravarthi et al. (2021). This dataset is also HASOC-Offensive Language identification-DravidianCodeMix- 2021.

The major objective of these shared tasks was to serve as a testbed for evaluating different techniques, allowing researchers to understand better how offensive language is transmitted on social media in Dravidian languages. These tasks have been highly successful, attracting wide interest at FIRE HASOC-Offensive Language Identification-DravidianCodeMix-2020 and DravidianLangTech-2021: they were one of the most popular FIRE tasks in 2020, attracting 96 participating teams, for DravidianLangTech-2021 attracting 129 participants.

We first introduce the datasets used in the offensive language identification shared tasks for Dravidian languages (Section 3). We present a detailed description of the dataset generation, with examples and inter-annotator agreement. Next, we describe the shared tasks, train test splits, and the evaluation setup in Section 4. In Section 5, we present the results across the shared task and discuss the systems presented by top 3 teams for all tasks. Finally, we compare the shared task to other comparable initiatives and suggest future research options.

## 2. Dravidian Languages

Dravidian languages is the name used to describe the 26 languages spoken in South India (Caldwell 1856), which are split into four groups: 11 in the Southern group, 7 in the South-Central group, 5 in the Central group, and 3 in the Northern group. Except for the three languages (Tamil, Malayalam, and Kannada) considered for this research along with Telugu, many of the 26 Dravidian languages are non-literary. Tamil, Malayalam and Kannada are under the subgroup of Dravidian South, while Telugu belongs to the Dravidian subgroup of the South Dravidian Vikram and Urs (2007) subgroup. Nonliterary languages are mostly used by indigenous minority people. The four literary languages are widely used in modern culture in literature, public communications, government institutions, academic settings, and many other locations in an ordinary person's day-to-day existence.

**Table 1.** Examples from the dataset used in the HASOC 1 shared task. The dataset included Malayalam Youtube comments in code-mixed, English transliterated, and native Malayalam scripts. We present the type of YouTube comment, the original script, the corresponding English translation, and the label.

|   | Type | Example | English translation | Label |
|---|------|---------|---------------------|-------|
| 1 | code-mixed | Oh my God ഇതു ചീറും | Oh my God, this will be a blast! | Not Offensive |
| 2 | English transliteration | Ella oollapadathinteyum stiram cheruva. 8 nilayil padam pottum. | Exact ingredient of all flop movies. This movie is going to be a failure. | Offensive |
| 3 | Mal. native script | പാവം സഞ്ജീവ് പിള്ളേയ്ക്കു എന്തെങ്കിലും ക്രെഡിറ്റ് | Please give some credits to Sanjeev pillai also you assholes. | Offensive |
| 4 | Mal. native script | കൊടുക്കെടാ നായിം നറമക്കെള രാക്ഷസൻ ഐമകൾനോടികൾ ഒെക്ക അങ്ങോട്ട് മാറിനിക്ക് ഇനി ഇവിടെ 5 പാതിര ഭരിക്കും | Marini will now rule the 5th floor here. | Not offensive |

The morphology of the Dravidian language is agglutinating and solely suffixal. Words are made up of morphemes, which are tiny components. Stems and affixes are the two main types of morphemes. Words are formed up of morphemes that are concatenated according to the language's grammar. Each Dravidian language have its own script (Krishnamurti 2003; Sakuntharaj and Mahesan 2016 2017). The Dravidian languages scripts were first documented around 580 BCE on pottery in Tamili script [a] from the districts of Keezhadi, Sivagangai, and Madurai in Tamil Nadu, India (Sivanantham and Seran 2019)[b]. Although the languages have their own scripts, social media users often use the Latin script for typing in these languages due to its ease of use and accessibility in handheld devices and computers.

## 3. Dataset Description

This section explains how we gathered and annotated our datasets of brief social media comments. We begin by explaining the HASOC-Dravidian 2020 data, which includes both tweets and Youtube comments. We then explain the data used at the shared task at DravidianLangTech 2021, and this dataset was built on Youtube comments.

### 3.1 HASOC-Dravidian 2020 data

There were two tasks associated with HASOC-Dravidian 2020, which were conducted on three datasets. Task 1 focused on Malayalam, and Task 2 focused on both Malayalam and Tamil. The Malayalam dataset was built only on Youtube comments, while the Tamil dataset was built on YouTube comments and comments from the Helo App.

For the dataset used in Task 1, we downloaded data from YouTube comments. The comments were downloaded from movie trailers during 2019 using the YouTube comment scraper tool [c]. We downloaded all comments for a given video using this tool. The comments included text in Malayalam Script, transliterated Malayalam text in Latin script, or a mix of Malayalam and Latin script. We observed code-mixing at word level, intra-sentential level and inter-sential level (Chakravarthi et al, 2020a) in our dataset. We built a dataset on these comments, where the annotators identified if the YouTube comment is offensive or not. Then, we used google forms to annotate the comments where each comment was annotated at least by two annotators who were proficient in Malayalam.

---
[a] also called Damili or Dramili or Tamil-Brahmi

**Table 2.** Examples from the dataset used in the HASOC 2 shared task. The dataset included both Malayalam and Tamil instances. The Malayalam dataset was built on Youtube comments and the Tamil dataset was built on Helo App. comments. Both Malayalam and Tamil datasets only included English transliterated instances. That is, they did not contain code-mixed or native Malayalam or Tamil scripts. We present the language of the instance, the original script, the corresponding English translation, and the label.

|   | Lang. | Example | English translation | Label |
|---|-------|---------|---------------------|-------|
| 1 | Mal. | Ninnae njan kollumeda thendi. | I will kill you bastard. | Offensive |
| 2 | Mal. | Athu thaniku hip hopum rapum thamilula difference ariyathonda. | There is no difference between hip hop and rap. | Not offensive |
| 3 | Tam. | Andha dark small beans eppadi cook pannuvinga. I tried. But not cooking properly. | How to cook those dark small beans. I tried. But not cooking properly. | Not Offensive |
| 4 | Tam. | Take it this thevidiya Kandipa indha page admin Oru Mutual Punda Vijay fan ha Dhan erupan. | Take it this whore, this page admin must be on of the foolish pussy Vijay fan. | Offensive |

We calculated Cohen's Kappa on of the dataset and found the agreement at 0.82. Based on McHugh (2012) 0.82 is substantial and shows that level of agreement is strong and 64-81% data are reliable. We present examples for task 1 in Table 1. In example 1, the comment is code-mixed in Malayalam script and English. The comment describes about the excitement of a fan who is hopeful of the movie being a blast. This comment hence clearly is *Not offensive*. In example 2, the comment is written entirely in the Latin script, and it is an *Offensive* comment. Though not very explicit, the author of the comment says that the movie will be a flop without strong reason expressing his hate. Example 3 represents a case when all the comments are written entirely in Malayalam script, and it is *Offensive*. The presence of a swear word in the comment makes it *Offensive*. Finally, in example 4, the comment is *Not offensive* and is entirely written in the Malayalam script. The dataset used for Task 2 included both Tamil and Malayalam instances. First, we collected YouTube comments for the Malayalam dataset entirely in the Latin script. Then, we built a dataset on these YouTube comments similar to the process in task 1. In the case of the Tamil dataset, we collected tweets and comments from the Helo App. Here, we considered instances that used only Latin characters. All the instances in the Tamil data did not use the Tamil script but only the Latin script. Then, we built a dataset on these tweets and comments from Help App, where the annotators identified if it is offensive or not offensive. Two Tamil speaking annotators did annotation. We used google forms to annotate the comments where each comment was annotated at least by two annotators.

**Table 3.** Corpus statistics of the datasets used in the shared task on HASOC- Offensive Language Identification in Dravidian languages at the FIRE 2020.

|  | Task1-Mal | Task2-Mal | Task2-Tam |
|---|---|---|---|
| Number of words | 287,299 | 415,109 | 591,841 |
| Vocabulary size | 9,292 | 24,834 | 26,008 |
| Number of comments | 4,000 | 5,000 | 4,940 |
| Number of sentences | 5,492 | 5,064 | 5,296 |
| Average number of words per sentence | 9 | 11 | 18 |
| Average number of sentences per comment | 1 | 1 | 1 |

This dataset included 5,000 Malayalam instances and 4,940 Tamil instances. We present examples for task 2 in Table 2. We calculated Cohen's Kappa on of the dataset and found the agreement at 0.69 for Malayalam and 0.73 for Tamil. Based on McHugh (2012) .60–.79 is substantial and shows that level of agreement is moderate. In example 1, a Malayalam sentence is written using the Latin scripts, and it threatening and insulting, so it is annotated as offensive. In example 2, Malayalam non-offensive comment was given in the Latin script. Example 3 represents a case when a Tamil non-offensive comment is written in the Latin script about cooking. Finally, in example 4 offensive Tamil comment is written in the Latin script.

Table 4 presents the corpus statistics of the dataset used in the HASOC task 1 and task 2. In case of the HASOC task 1, there are 4,000 Malayalam instances with 287,299 words and 9,292 unique words. There are a total of 5,492 sentences. On an average each word sentece had 9 words. In case of the HASOC task 2, there are 5,000 Malayalam instances and 4,940 Tamil comments. There are 415,109 words in the Malayalam dataset with 49,781 unique words. There are a total of 5, 064 sentences. The average number of words per sentence is 11 and the average number of sentences per comment is one. On an average each word sentece had 11 words. Regarding the Tamil dataset there are 591,841 words, of which 26,008 are unique. The dataset includes a total of 5,296 sentences. On an average each word sentece had 18 words.

**Table 4.** Label distribution across the datasets used in the HASOC taks. In case of the HASOC 1 Malayalam task, the label distribution is uneven with less instances in the *Offensive* class. Regarding HASOC 2 tasks, the label distribution between *Offensive* and *Not offensive* classes are almost even. (49% vs. 51% and 50% vs. 50%)

| Class | Task1-Mal. | % | Task2-Mal. | % | Task2-Tam. | % |
|---|---|---|---|---|---|---|
| Offensive | 705 | 18 | 2,465 | 49 | 2,455 | 50 |
| Not Offensive | 3,295 | 82 | 2,535 | 51 | 2,485 | 50 |
| Total | 4,000 | 100 | 5,000 | 100 | 4,940 | 100 |

We present the label distribution of the HASOC tasks in Table 4. In case of the HASOC 1 task where only Malayalam instances are present, there are 705 *Offensive* instances and 3295 *Not offensive instances*. The percentage distribution is 18% vs. 82%, where the *Offensive* class is the minority. In case of the HASOC 2 tasks, the percentage distribution is almost 50-50 for both the Malayalam and the Tamil tasks.

### 3.2 DravidianLangTech Data

The dataset used in the Offensive language identification shared task at the DravidianLangTech workshop included three languages Malayalam, Tamil, and Kannada. Data was compiled from different film trailers of Malayalam, Tamil and Kannada languages from YouTube comments in 2019.

First we used the *YouTube Comment Scraper tool*[d] to collect comments from YouTube social media. Next, we gathered comments in all three languages that had code-mixing at various levels of the text, with enough representation for each class. We used the *Langdetect library* [e] to detect the language of intent (Malayalam, Tamil or Kannada) and removed all comments that are not in the intended language. Due to the fact that our data was gathered via social media, it comprises many forms of real-time code-mixing. The dataset includes all types of code-mixing, from fully monolingual texts in native languages through script, word, and morphological mixing, as well as inter-sentential and intra-sentential changes. We used these comments to generate the dataset by annotating for each language. Annotations were done on google forms, where each comment is annotated by at least three annotators proficient in the intended language. We added a new label *Not in intended language* as some instances were not in the intended language but were wrongly identified by the *Langdetect library*. Unlike the HASOC tasks, we did not only annotate if a given comment is *offensive* or *not-offensive*, but also presented fine-grained labels indicating if the offense is targeted to a group or individual following Zampieri et al. (2019). To this extent, we present 6 labels for this task. We describe the labels below.

---

[d]https://github.com/philbot9/youtube-remarkscraper

**Table 5.** Examples from the dataset used in the shared task on Offensive Language Identification in Dravidian languages at the First Workshop on Speech and Language Technologies for Dravidian Languages (DravidianLangTech-2021). The dataset included Malayalam, Tamil, and Kannada Youtube comments in code-mixed, English transliterated, and the corresponding native scripts. We present the language of the Youtube comment, type of the comment, the original script, the corresponding English translation, and the label. Unlike the binary-labelled instances in the HASOC tasks, the instances in this task were given one of the six labels described in Section 3.2.

| | Lang. | Type | Example | English translation | Label |
|---|---|---|---|---|---|
| 1 | Tam. | Tamil script | படம் ரொம்ப நல்ல இருந்தது | The movie was very good | Not offensive |
| 2 | Mal. | code-mixed | Nallatrailernu okke keridislikeadikunne ethelum thanthayillathavanmar aayirikum. poyi chavinedey... | Those who dislike any trailers will probably be assholes. Go to hell... | Offensive Untargeted |
| 3 | Tam. | English transliteration | Ghibran p—a vitta ungalukku vera aale kedaika laya da. | Dont you get anyone else other than Ghibran V—a | Offensive Targeted Individual |
| 4 | Kan. | code-mixed | Guru ee desha uddhara agalla bedu bhai indian youth waste bedu | Brother this country won't develop as Indian youth are waste | Offensive-Targeted-Insult-Group |
| 5 | Mal. | code-mixed | pavam sanjeev pillaiku enthengilum credic kodukeda nayinte makkale | please give some credits to Sanjeev pillai also you assholes | Offensive untargeted |
| 6 | Kan. | code-mixed | Finally, sonu gowda b day dhinane tiktok ban aythu | Finally, TikTokgot banned on the birthday of fucker Sonu Gowda | Offensive-Targeted-Insult-Individual |

- **Not Offensive**: Comment/post does not have offence, obscenity, swearing, or profanity.
- **Offensive Untargeted**: Comment/post have offence, obscenity, swearing, or profanity not directed towards any target. These are the comments/posts which have inadmissible language without targeting anyone.

---

[e]https://pypi.org/venture/langdetect/

**Table 6.** Corpus statistics of the datasets used in the shared task on Offensive Language Identification in Dravidian languages at the First Workshop on Speech and Language Technologies for Dravidian Languages (DravidianLangTech-2021).

| Language | Tamil | Malayalam | Kannada |
| --- | --- | --- | --- |
| Number of words | 511,734 | 202,134 | 65,702 |
| Vocabulary size | 94,772 | 40,729 | 20,796 |
| Number of comments | 43,919 | 20,010 | 7,772 |
| Number of sentences | 52,617 | 23,652 | 8,586 |
| Average number of words per sentence | 11 | 10 | 8 |
| Average number of sentences per comment | 1 | 1 | 1 |

- **Offensive Targeted Individual**: Comment/post have offence, obscenity, swearing, or profanity which targets an individual.
- **Offensive Targeted Group**: Comment/post have offence, obscenity, swearing, or profanity which targets a group or a community.
- **Offensive Targeted Other**: Comment/post have offence, obscenity, swearing, or profanity which does not belong to any of the previous two classes.
- **Not in indented language**: If the comment is not in the intended language. For example, in the Malayalam task, if the sentence does not contain Malayalam written in Malayalam script or Latin script, then it is not Malayalam.

We present examples from each language along with their type, English translation, and labels in Table 5.

In Example 1, the comment says that the movie was very good in the native Tamil script. Thus the label is *Not offensive*. In Example 2, the comment intends to offend everyone who dislikes the trailer, calling them a\*\*holes. The comment is a mix of English transliterated Malayalam and English, and annotators chose the label *Offensive Untargeted*. Example 3 is a case of the *Offensive Targeted Individual* label, where the comment specifically offends an individual. Here the comment targets to offend a famous Tamil musician Ghibran. In the case of example 4, the comment is in Kannada written in the Latin script. The comment insults Indian youth calling them a waste, so it is *Offensive Targeted Insult Group*. In Example 5, Malayalam offensive untargeted comment is written in the Latin script. Finally, example 6 indicates a case of targeting a person by name; this is a Kannada comment written in the Latin script *Offensive Targeted Insult Individual*.

We present corpus statistics of the dataset used in the shared task at the DravidianLangTech workshop in Table 6. Because of the nature of our annotation system, we used Krippendorff's alpha. Krippendorff's alpha allows for partial data and hence does not require every annotator to annotate every comment. We achieved 0.74, 0.83, 0.84 for Tamil, Malayalam, and Kannada, respectively. From Table 6, we can see that vocabulary size is massive for Tamil and Malayalam due to the code-mixing and complex morphology of these languages.

Table 7 presents the label distribution of the DravidianLangTech-2021 Malayalam, Tamil and Kannada datasets. In all the datasets the label *Not offensive* dominates. This is expected and in par with the real world. Malayalam dataset has the most *Not offensive* followed by Tamil (73%) and Kannada (56%). The number of *Offensive* labels which are spread across the labels *Offensive Untargeted*, *Offensive Targeted Individual*, *Offensive Targeted Group*, and *Offensive Targeted Others* are also relatively smaller than other languages. This means only 3% of the entire Malyalam dataset are offensive. In case of the Tamil data, we identify several offensive instances spread across the different offensive categories. Only 4% of the entire Tamil dataset has instances that are not in Tamil. Regarding Kannada, 25% of the entire dataset is not in Kannada. This indicates the inefficiency of the Langdetect library we used to identify the language of a given comment. Additionally around 18% of the entire dataset indicates hate speech.

**Table 7.** Label distribution across the entire dataset for the datasets used in the shared task on Offensive Language Identification in Dravidian languages at the First Workshop on Speech and Language Technologies for Dravidian Languages (DravidianLangTech-2021). Recall that unlike the HASOC tasks, the labels are not binary.

| Class | Malayalam | % | Tamil | % | Kannada | % |
|---|---|---|---|---|---|---|
| Not Offensive | 17,697 | 89 | 31,808 | 73 | 4,336 | 56 |
| Offensive Untargeted | 240 | 1 | 3,630 | 8 | 278 | 3 |
| Offensive Targeted Individual | 290 | 1 | 2,965 | 7 | 628 | 8 |
| Offensive Targeted Group | 176 | 1 | 3,140 | 7 | 418 | 5 |
| Offensive Targeted Others | - | 0 | 590 | 1 | 153 | 2 |
| Not in indented language | 1,607 | 8 | 1,786 | 4 | 1,898 | 25 |
| Total | 20,010 | 100 | 43,919 | 100 | 7,772 | 100 |

## 4. Task Descriptions

In this section, we first present a detailed description of the HASOC shared tasks conducted in 2020. Later, we describe the DravidianLangTech-2021 offensive language identification task. We begin with briefly describing the goal of the tasks, the distribution of labels in the train-test splits and the task procedure. Both tasks followed a two-phase procedure, where the training data was released in the first phase, and the labels were evaluated in the next phase.

### 4.1 HASOC Offensive Language Identification- FIRE 2020

The goal of this task is to identify offensive language from a code-mixed dataset of comments/posts in Dravidian Languages (Malayalam, Malayalam-English, and Tamil-English) collected from social media. The comment/post may contain more than one sentence but the average sentence length of the corpora is 1. Each comment/post is annotated with offensive language label at the comment/post level. The task-1 dataset also has class imbalance problems depicting real-world scenarios. The participants were provided with development, training and test dataset.

- **Task1:** This is a message-level label classification task. Given a YouTube comment in code-mixed Malayalam, systems have to classify it into offensive or not-offensive.
- **Task2:** This is a message-level label classification task. Given a tweet or Youtube comments in Tanglish and Manglish (Tamil and Malayalam using written using Roman Characters), systems have to classify it into offensive or not-offensive.

**Table 8.** Label distribution across train test splits of the dataset in HASOC task 2. The labels are almost evenly split between the two classes *Offensive* and *Not offensive*. Recall that HASOC task 1 includes instances only from Malayalam.

|  |  | Train | % | Test | % |
|---|---|---|---|---|---|
| Mal | Offensive | 639 | 18 | 66 | 17 |
|  | Not offensive | 2,961 | 82 | 334 | 83 |

We present the label distribution across the train test splits for HASOC task 1 in Table 8.

**Table 9.** Label distribution across train test splits of the dataset in HASOC task 2. The labels are almost evenly split between the two classes *Offensive* and *Not offensive* for both the languages Malayalam and Tamil.

|     |               | Train | %  | Test | %  |
|-----|---------------|-------|----|------|----|
| Mal | Offensive     | 1,953 | 49 | 512  | 51 |
|     | Not offensive | 2,047 | 51 | 488  | 49 |
| Tam | Offensive     | 1,980 | 49 | 475  | 51 |
|     | Not offensive | 2,020 | 51 | 465  | 49 |

We present the label distribution across the train test splits for HASOC task 2 in Table 9. Unlike the HASOC 1 Malayalam task, the label distribution for HASOC 2 task is evenly spread across both the train and test set.

### 4.2 DravidianLangTech Offensive Language Identification 2021

Offensive language identification for Dravidian languages at different levels of complexity was developed following the work of (Zampieri et al, 2019). It was customized to our annotation method from a three-level hierarchical annotation schema. A new category *Not in intended language* was added to include comments written in a language other than the Dravidian languages. Annotations decisions for offensive language categories were split into six labels to simplify the annotation process.

**Table 10.** Train-Development-Test Data Distribution with 90%-5%-5% train-dev-test for Offensive Language Identification

|             | Tamil  | Malayalam | Kannada |
|-------------|--------|-----------|---------|
| Training    | 35,139 | 16,010    | 6,217   |
| Development | 4,388  | 1,999     | 777     |
| Test        | 4,392  | 2,001     | 778     |
| Total       | 43,919 | 20,010    | 7,772   |

Unlike the HASOC tasks, we provide the participants with the development data and split the dataset into 90%-5%-5% for training, development, and testing respectively. Table 10 presents the distribution of the data instances across the training, development, and test sets. We ensured that the labels follow a similar distribution as observed in the entire dataset.

### 4.3 Phase 1: Release training data

In case of both the HASOC tasks and the Dravidian Langtech shared task, we first released the training dataset with labels. We released the dataset in CSV format in CodaLab. CodaLab is an established platform to organize shared-tasks. Participants made use of this dataset to build models and evaluated on their own development sets, but choosing a portion of data from the training set. Later, participants were given the test set without any labels and asked their predictions in the CSV format. We followed the evaluation procedure as described in 4.4 to generate the leader board.

### 4.4 Phase 2: Evaluation Setup

In this phase we evaluated the participants' predictions. Each participating team submitted their generated prediction for evaluation. Predictions are submitted via Google form to the organizing committer to evaluate the systems. Initially, we intended to use CodaLab for our evaluating the submitted models. However, we faced issues with running evaluation, and so we chose to evaluate the predictions manually. We used weighted average F1 score as our evaluation metric.

In both HASOC and Dravidian Langtech, all teams were allowed a total of 3 submission per Task. Participants could submit predictions based on the language of their choosing. That is, in case of HASOC task 1, participants did not have a choice as it only used the Malayalam dataset. In case of HASOC task 2, participants were allowed to submit their predictions either for Malayalam or Tamil or both. Each participant could submit up to three sets of predictions per language. In case of Dravidian Langtech shared task, participants were allowed to submit their predictions either for Malayalam, Tamil, Kannada, or all of them. Similar to the HASOC tasks, participants could submit up to three sets of predictions per language. The systems were evaluated on precision, recall and F1-score. We used the weighted average F1 score to rank the participants. This takes into account the varying degrees of importance of each class in the dataset. We used a classification report tool from Scikit learn[1]. We present the formulae for Precision, Recall, F1 score, weighted precision, weighted Recall, and weighted F1 score below.

$$\text{Precision} = \frac{TP}{TP + FP} \quad (1)$$

$$\text{Recall} = \frac{TP}{TP + FN} \quad (2)$$

$$\text{F-Score} = 2 * \frac{\text{Precision} * \text{Recall}}{\text{Precision} + \text{Recall}} \quad (3)$$

$$P_{\text{weighted}} = \sum_{i=1}^{L}(P \text{ of } i \times \text{ Weight of } i) \quad (4)$$

$$R_{\text{weighted}} = \sum_{i=1}^{L}(R \text{ of } i \times \text{ Weight of } i) \quad (5)$$

$$F1_{\text{weighted}} = \sum_{i=1}^{L}(F1 \text{ of } i \times \text{ Weight of } i) \quad (6)$$

## 5. Participants Results and Analysis

In this section we present and discuss the results of the offensive language identification systems built by the participants of the shared task in HASOC 2020 and Dravidian langtech. We discuss the top 3 best performing offensive language identification models per language for each task.

### 5.1 Results from HASOC Dravidian 2020

We present the results of the two subtasks in HASOC Dravidian 2020. In HASOC subtask 1, we work only with Malayalam dataset and present the results in Table 1 1.

---

[1] https://scikit-learn.org/stable/modules/generated/sklearn.metrics.classification_report.html

**Table 11.** Results of HASOC Task 1: Malayalam. The performance of the top 3 teams in the leader board are very similar in terms of Precision, Recall, and F-Score.

| TeamName | Precision | Recall | F-Score | Rank |
| --- | --- | --- | --- | --- |
| SivaSai@BITS (Sai and Sharma 2020) | 0.95 | 0.95 | 0.95 | 1 |
| IIITG-ADBU (Baruah et al. 2020) | 0.95 | 0.95 | 0.95 | 1 |
| CFILT-IITBOMBAY | 0.94 | 0.94 | 0.94 | 2 |
| SSNCSE-NLP (Bharathi and Silvia A 2021) | 0.94 | 0.94 | 0.94 | 2 |
| CENMates (Veena et al. 2020) | 0.93 | 0.93 | 0.93 | 3 |
| NIT-AI-NLP (Kumar and Singh 2020) | 0.93 | 0.93 | 0.93 | 3 |
| YUN (Dong 2020) | 0.93 | 0.93 | 0.93 | 3 |
| Zyy1510 (Zhu and Zhou 2020) | 0.93 | 0.93 | 0.93 | 3 |
| Gauravarora (Arora 2020) | 0.92 | 0.91 | 0.91 | 4 |
| WLV-RIT (Ranasinghe and Zampieri 2020) | 0.89 | 0.90 | 0.89 | 5 |
| Kjdong( only not) | 0.70 | 0.83 | 0.76 | 6 |
| Ajees (Ajees 2020) | 0.69 | 0.38 | 0.44 | 7 |

**HASOC Task 1: Malayalam.** The results of the task HASOC task 1: Malayalam are presented in Table 11. The first and second position teams utilized transformer-based model for classification of YouTube comments into offensive and Non Offensive. SivaSai@BITS and IIITG-ADBU ranked one with an F1-score of 0.95 with precision and recall score of 0.95. Both SivaSai@BITS and IIITG-ADBU used transformer based model XMLRoberta, SVM with XMLRoberta respectively. Teams CFILT-IITBOMBAY and SSNCSE-NLP achieved the second position with an F-score of 0.94 adapting BERT model with ML, BERT model with a novel training strategy of transliteration of Romanized text into the native script and found to be effective. The top four teams attained F-score higher than 0.90. The difference between the evaluation scores of top teams is minimal. Another important fact visible from the results is the Support Vector Machine classifiers with TF-IDF features also reach top positions. Other systems submitted to the task use deep learning models using Bidirectional LSTM, LSTM, CNN and ULMFit.

**HASOC Task 2: Malayalam and Tamil**

In this task, we work with datasets from both Malayalam and Tamil. We present the results of Malayalam dataset in Table 12. The precision and recall Top 3 systems are equal and above 90% indicating the equal number of false positive and false negative predictions by the systems.

**HASOC Task 2: Malayalam.** The results of this task are presented in Table 12. Here, CENmates reached the first position with an F-score of 0.78. Teams SivaSai@BITS and KBCNMUJAL bagged the second place, and their F-score was 0.77. The scores of the top five teams are close. Team CENmates used TF-IDF features with character n-gram as features for classification using machine learning algorithms. SivaSai@BITS used transformer based model XMLRoberta for this task. Team KBCNMUJAL used character and word n-grams features for with machine learning classifiers. Other teams used transformer-based models and Deep Learning-based models. Among top 8 team, except the topmost team, recall of all the other teams is comparatively lesser than precision. Precision indicates how precise the model is out of those predicted positive whereas, recall finds how many of the actual positives our model capture through labeling it as Positive. Hence in this sub task, most of the models are precise among those predicted positive.

**Table 12.** Results of HASOC Task 2: Malayalam. The overall performance is less than that of HASOC task 1, despite both these tasks including Malayalam instances.

| TeamName | Precision | Recall | F-Score | Rank |
| --- | --- | --- | --- | --- |
| CENmates (Veena et al, 2020) | 0.78 | 0.78 | 0.78 | 1 |
| SivaSai (Sai and Sharma, 2020) | 0.79 | 0.75 | 0.77 | 2 |
| KBCNMUJAL (Pathak et al, 2020) | 0.77 | 0.77 | 0.77 | 2 |
| IIITG-ABDU (Baruah et al, 2020) | 0.77 | 0.76 | 0.76 | 3 |
| SSNCSE-NLP (Bharathi and Silvia A, 2021) | 0.78 | 0.74 | 0.75 | 4 |
| Gauravarora (Arora, 2020) | 0.76 | 0.72 | 0.74 | 5 |
| CFILT (Singh and Bhattacharyya, 2020) | 0.74 | 0.70 | 0.72 | 6 |
| NITP (Kumar and Singh, 2020) | 0.71 | 0.68 | 0.69 | 7 |
| Ajees (Ajees, 2020) | 0.72 | 0.67 | 0.68 | 8 |
| Baseline | 0.69 | 0.68 | 0.68 | 8 |
| YUN (Dong, 2020) | 0.67 | 0.67 | 0.67 | 9 |
| Zyy1510 (Zhu and Zhou, 2020) | 0.68 | 0.67 | 0.67 | 9 |
| CUSAT (Renjit, 2020) | 0.54 | 0.54 | 0.54 | 10 |

**Table 13.** Results of HASOC Task 2: Tamil. Interestingly, the overall results of this task is better than that of HASOC task 2: Malayalam, despite both these tasks including English transliterated instances in Malayalam or Tamil.

| TeamName | Precision | Recall | F-Score | Rank |
| --- | --- | --- | --- | --- |
| SivaSaiBITS (Sai and Sharma, 2020) | 0.90 | 0.90 | 0.90 | 1 |
| SSNCSE-NLP (Bharathi and Silvia A, 2021) | 0.88 | 0.88 | 0.88 | 2 |
| Gauravarora (Arora, 2020) | 0.88 | 0.88 | 0.88 | 2 |
| KBCNMUJAL (Pathak et al, 2020) | 0.87 | 0.87 | 0.87 | 3 |
| IIITG-ADBU (Baruah et al, 2020) | 0.87 | 0.87 | 0.87 | 3 |
| Zyy1510 (Zhu and Zhou, 2020) | 0.88 | 0.87 | 0.87 | 3 |
| CENmates (Veena et al, 2020) | 0.86 | 0.86 | 0.86 | 4 |
| CFILT (Singh and Bhattacharyya, 2020) | 0.86 | 0.86 | 0.86 | 4 |
| YUN (Dong, 2020) | 0.85 | 0.85 | 0.85 | 5 |
| NIT-AI-NLP (Kumar and Singh, 2020) | 0.84 | 0.84 | 0.84 | 6 |
| Baseline | 0.85 | 0.84 | 0.84 | 6 |
| Ajees (Ajees, 2020) | 0.84 | 0.83 | 0.83 | 7 |

**HASOC Task 2: Tamil.** The results of this task are presented in Table 13. Here, team SivaSai@BITS placed in the first position with an F-score of 0.90 with transformer-based model. Team SSNCSE-NLP scored an F-score of 0.88 and obtained second position using BERT in combination with machine learning models. Gauravarora adopted a pre-trained ULMFiT for the classification also grabbed the second position. Three teams reached third position with an F-score of 0.87 adopting different features and classifiers for the prediction task. Team KBCNMUJAL uses character n-gram and word ngram features for representing text with ensemble models for classification. Team IIITG-ADBU used SVM and XLM-RoBERTa based classifiers. Team Zyy1510 used an ensemble of BiLSTM, LSTM+Convolution and a Convolution for the classification of social media texts into OFF and NOT. In this task the precision and recall of most of the teams are equal except the baseline model indicating the equal number of false positive and false negative predictions by the systems.

### 5.2 Results from Dravidian langtech

In this shared task, we worked with three datasets corresponding to the languages Malayalam, Tamil, and Kannada. We present the results for each language and discuss the models submitted by the top performing teams.

**Dravidian Langtech: Malayalam.** We present the results of Dravidian Langtech: Malayalam in Table 14. The teams *hate-alert*, (Balouchzahi et al. 2021), *indicnlp@kgp*, and *bitions* shared the first position with an F1 Score of 0.97. Team *hate-alert* (Saha et al. 2021) used multilingual BERT (mBERT), XLM Roberta, Indic BERT, and MURIL Fusion models combining CNN and BERT models. MUCS (Balouchzahi et al. 2021) used two classification models namely, COOLI-Ensemble models and COOLI-Keras models. Team *indicnlp@kgp* (Kedia and Nandy 2021) used various machine learning models including Random Forest, Naïve Bayes and Linear SVM . Linear SVM results are better compared to their other models. Team *bitions* (Tula et al. 2021) have used Indic BERT, DistilmBERT and ULMFit models . The teams *hypers* (Vasantharajan and Thayasivam 2021), *OFFLangOne* (Dowlagar and Mamidi 2021), and *SJ-AJ* (Jayanthi and Gupta 2021) who came in the second position used variations of BERT models. The teams *CUSATNLP* (Renjit and Idicula 2021), *IIITK* (Ghanghor et al. 2021), and *IRNLP-DAIICT* (Dave et al. 2021) also used variations of BERT models including mBERT and XLM Roberta and were positioned at rank 3. Interestingly, *SSNCSE-NLP* (Bharathi and Silvia A 2021) used traditional algorithms such as K-Nearest Neighbours, Support Vector Classifier and Multi Layered Perceptron with TF-IDF and Bag-of-words features ranking at position 3.

**Dravidian Langtech: Tamil.** We present the results of Dravidian Langtech: Tamil in Table 15. Team *hate-alert* won the position followed by *indicnlp@kgp* at the second position. The third position was shared by *ZYJ123*, *ALIB2BAI*, *SJAJ*, *No offense*, and *NLP@CUET*. The systems by *hate-alert* and *indicnlp@kgp* are similar to what they built for the Malayalam model.

*ZYJ123* (Zhao 2021) proposed a system based on the multilingual model XLM-Roberta and DPCNN. *SJ-AJ* (Jayanthi and Gupta 2021) system is an ensemble of mBERT and XLM-RoBERTa models which leverage task-adaptive pre-training of multilingual BERT models with a masked language modeling objective. *No offense* (Awatramani 2021) used mBERT-cased and XLMRoBERTa. *NLP@CUET* (Sharif et al. 2021) employed two machine learning techniques (LR, SVM), three deep learning (LSTM, LSTM+Attention) techniques and three transformers (m-BERT, Indic-BERT, XLM-R) based methods. They showed that XLM-R outperforms other techniques.

**Dravidian Langtech: Kannada.** We present the results of Dravidian Langtech: Kannada in Table 16. When compared to that of Malayalam and Tamil offensive language task, the accuracy achieved by even the Top 3 teams for Kannada is very low. This phenomena is attributed to the less number of training instances in the dataset. Also, the class imbalance problem that is inherent in the dataset makes it worse. *SJ_AJ* reach the first position, followed by *hate_alert* in the second position. We have discussed these two systems previously in the Malayalam and Tamil task. The third position is shared by *indicnlp@kgp*, *Codewithzichao*, and *IIITK* . We have already discussed *indicnlp@kgp*. The team used Multilingual Transformers, and the team *IITK* used the used multilingual BERT-base model

Table 14. Results of DravidianLangTech Offensive Language Identification 2021: Malayalam

| Team-Name | Precision | Recall | F1 Score | Rank |
| --- | --- | --- | --- | --- |
| hate-alert (Saha et al. 2021) | 0.97 | 0.97 | 0.97 | 1 |
| MUCS (Balouchzahi et al. 2021) | 0.97 | 0.97 | 0.97 | 1 |
| indicnlp@kgp (Kedia and Nandy 2021) | 0.97 | 0.97 | 0.97 | 1 |
| bitions (Tula et al. 2021) | 0.97 | 0.97 | 0.97 | 1 |
| hypers (Vasantharajan and Thayasivam 2021) | 0.96 | 0.96 | 0.96 | 2 |
| ALI-B2B-AI | 0.96 | 0.97 | 0.96 | 2 |
| OFFLangOne (Dowlagar and Mamidi 2021) | 0.96 | 0.96 | 0.96 | 2 |
| SJ-AJ (Jayanthi and Gupta 2021) | 0.97 | 0.97 | 0.96 | 2 |
| CUSATNLP (Renjit and Idicula 2021) | 0.95 | 0.95 | 0.95 | 3 |
| IIITK (Ghanghor et al. 2021) | 0.94 | 0.95 | 0.95 | 3 |
| SSNCSE-NLP (Bharathi and Silvia A 2021) | 0.95 | 0.96 | 0.95 | 3 |
| IRNLP-DAIICT (Dave et al. 2021) | 0.96 | 0.96 | 0.95 | 3 |
| Codewithzichao (Li 2021) | 0.93 | 0.94 | 0.94 | 4 |
| JudithJeyafreeda (Andrew 2021) | 0.94 | 0.94 | 0.93 | 5 |
| cs (Chen and Kong 2021) | 0.92 | 0.94 | 0.93 | 5 |
| Zeus | 0.92 | 0.95 | 0.93 | 5 |
| Maoqin (Yang 2021) | 0.91 | 0.94 | 0.93 | 5 |
| NLP@CUET (Sharif et al. 2021) | 0.92 | 0.94 | 0.93 | 5 |
| ZYJ123 (Zhao 2021) | 0.91 | 0.94 | 0.92 | 6 |
| hub (Huang and Bai 2021) | 0.89 | 0.93 | 0.91 | 7 |
| No offense (Awatramani 2021) | 0.9 | 0.93 | 0.91 | 7 |
| e8ijs | 0.89 | 0.92 | 0.9 | 8 |
| snehan-coursera | 0.9 | 0.89 | 0.89 | 9 |
| KU-NLP | 0.87 | 0.91 | 0.89 | 9 |
| KBCNMUJAL | 0.93 | 0.88 | 0.89 | 9 |
| IIITT (Yasaswini et al. 2021) | 0.84 | 0.87 | 0.86 | 10 |
| Agilna | 0.85 | 0.88 | 0.86 | 10 |
| Amrita-CEN-NLP | 0.9 | 0.82 | 0.85 | 11 |
| professionals (Sreelakshmi et al. 2021) | 0.89 | 0.84 | 0.85 | 11 |
| JUNLP (Nair and Fernandes 2021, Garain et al. 2021) | 0.77 | 0.43 | 0.54 | 12 |

## 6. Error Analysis

In this section, we analyze the common errors found in the top performing systems for the HASOC and Dravidian Langtech tasks. We begin with analyzing the errors in the HASOC task followed by the Dravidian Langtech task. We also target to present insights into the most common errors based on a qualitative analysis.

Table 15. Results of DravidianLangTech Offensive Language Identification 2021: Tamil

| Team-Name | Precision | Recall | F1 Score | Rank |
|---|---|---|---|---|
| hate-alert (Saha et al. 2021) | 0.78 | 0.78 | 0.78 | |
| indicnlp@kgp (Kedia and Nandy 2021) | 0.75 | 0.79 | 0.77 | |
| ZYJ123 (Zhao 2021) | 0.75 | 0.77 | 0.76 | |
| ALI-B2B-AI | 0.75 | 0.78 | 0.76 | |
| SJ-AJ (Jayanthi and Gupta 2021) | 0.75 | 0.79 | 0.76 | |
| No offense (Awatramani 2021) | 0.75 | 0.78 | 0.76 | |
| NLP@CUET (Sharif et al. 2021) | 0.75 | 0.78 | 0.76 | |
| Codewithzichao (Li 2021) | 0.74 | 0.77 | 0.75 | |
| hub (Huang and Bai 2021) | 0.73 | 0.78 | 0.75 | |
| MUCS (Balouchzahi et al. 2021) | 0.74 | 0.77 | 0.75 | |
| bitions (Tula et al. 2021) | 0.74 | 0.77 | 0.75 | |
| IIITK (Ghanghor et al. 2021) | 0.74 | 0.77 | 0.75 | |
| OFFLangOne (Dowlagar and Mamidi 2021) | 0.74 | 0.75 | 0.75 | |
| spartans | 0.74 | 0.78 | 0.75 | |
| cs (Chen and Kong 2021) | 0.74 | 0.75 | 0.74 | |
| Zeus | 0.72 | 0.78 | 0.73 | |
| hypers (Vasantharajan and Thayasivam 2021) | 0.71 | 0.76 | 0.73 | |
| SSNCSE-NLP (Bharathi and Silvia A 2021) | 0.74 | 0.73 | 0.73 | |
| JUNLP (Garain et al. 2021) | 0.71 | 0.74 | 0.72 | |
| snehan-coursera | 0.7 | 0.73 | 0.71 | |
| IIITT (Yasaswini et al. 2021) | 0.7 | 0.73 | 0.71 | |
| IRNLP-DAIICT (Dave et al. 2021) | 0.72 | 0.77 | 0.71 | |
| CUSATNLP (Renjit and Idicula 2021) | 0.67 | 0.71 | 0.69 | |
| e8ijs | 0.71 | 0.76 | 0.69 | |
| KU-NLP | 0.66 | 0.75 | 0.67 | |
| cognizyr | 0.68 | 0.74 | 0.66 | |
| Agilna | 0.63 | 0.72 | 0.65 | |
| DLRG | 0.57 | 0.74 | 0.64 | |
| Amrita-CEN-NLP | 0.64 | 0.62 | 0.62 | |
| KBCNMUJAL (Sreelakshmi et al. 2021) | 0.75 | 0.56 | 0.62 | |
| JudithJeyafreeda (Andrew 2021) | 0.54 | 0.73 | 0.61 | |

**Table 16.** Results of DravidianLangTech Offensive Language Identification 2021: Kannada

| Team-Name | Precision | Recall | F1 Score | Rank |
|---|---|---|---|---|
| SJ-AJ (Jayanthi and Gupta 2021) | 0.73 | 0.78 | 0.75 | 1 |
| hate-alert (Saha et al. 2021) | 0.76 | 0.76 | 0.74 | 2 |
| indicnlp@kgp(Kedia and Nandy 2021) | 0.71 | 0.74 | 0.72 | 3 |
| Codewithzichao (L 2021) | 0.70 | 0.75 | 0.72 | 3 |
| IIITK (Ghanghor et al. 2021) | 0.70 | 0.75 | 0.72 | 3 |
| e8ijs | 0.70 | 0.74 | 0.71 | 4 |
| NLP@CUET(Sharif et al. 2021) | 0.70 | 0.74 | 0.71 | 4 |
| ALI-B2B-AI | 0.70 | 0.73 | 0.71 | 4 |
| Zeus | 0.66 | 0.75 | 0.7 | 5 |
| hypers (Vasantharajan and Thayasivam 2021) | 0.69 | 0.72 | 0.7 | 5 |
| bitions (Tula et al. 2021) | 0.69 | 0.72 | 0.7 | 5 |
| SSNCSE-NLP (Bharathi and Silvia A 2021) | 0.71 | 0.74 | 0.7 | 5 |
| MUCS (Balouchzahi et al. 2021) | 0.68 | 0.72 | 0.69 | 6 |
| ZYJ123 (Zhao 2021) | 0.65 | 0.74 | 0.69 | 6 |
| MUM | 0.69 | 0.71 | 0.69 | 6 |
| JUNLP (Garain et al. 2021) | 0.62 | 0.71 | 0.66 | 7 |
| OFFLangOne (Dowlagar and Mamidi 2021) | 0.66 | 0.65 | 0.65 | 8 |
| cs (Chen and Kong 2021) | 0.64 | 0.67 | 0.64 | 9 |
| hub (Huang and Bai 2021) | 0.65 | 0.69 | 0.64 | 9 |
| No offense (Awatramani 2021) | 0.62 | 0.70 | 0.64 | 9 |
| IRNLP-DAIICT (Dave et al. 2021) | 0.69 | 0.69 | 0.64 | 9 |
| JudithJeyafreeda (Andrew 2021) | 0.66 | 0.67 | 0.63 | 10 |
| CUSATNLP (Renjit and Idicula 2021) | 0.62 | 0.63 | 0.62 | 11 |
| Agilna | 0.58 | 0.65 | 0.59 | 12 |
| Amrita-CEN-NLP (Sreelakshmi et al. 2021) | 0.65 | 0.54 | 0.58 | 13 |
| snehan-coursera | 0.56 | 0.61 | 0.54 | 14 |
| KBCNMUJAL | 0.72 | 0.48 | 0.52 | 15 |
| IIITT (Yasaswini et al. 2021) | 0.46 | 0.48 | 0.47 | 16 |
| Simon (Que et al. 2021) | 0.60 | 0.30 | 0.33 | 17 |

### 6.1 Quantitative error analysis

Figure 5.2 shows the results of analysis over the top 3 systems per subtask per language in the test set. In case of the HASOC task 1, where all instances belong to the Malayalam dataset, 42 instances are always correctly predicted by the top 3 systems. There is only one instance that is always predicted wrongly by the top 3 systems. There are 365 instances where all top 3 systems predicted correctly, except for one of the top 3 systems. Similarly there are 19 instances where all top 3 systems predicted wrongly, except for one of the top 3 system. Recall that there are 400 instances in the test set of HASOC 1 Malayalam dataset as in Table 8.

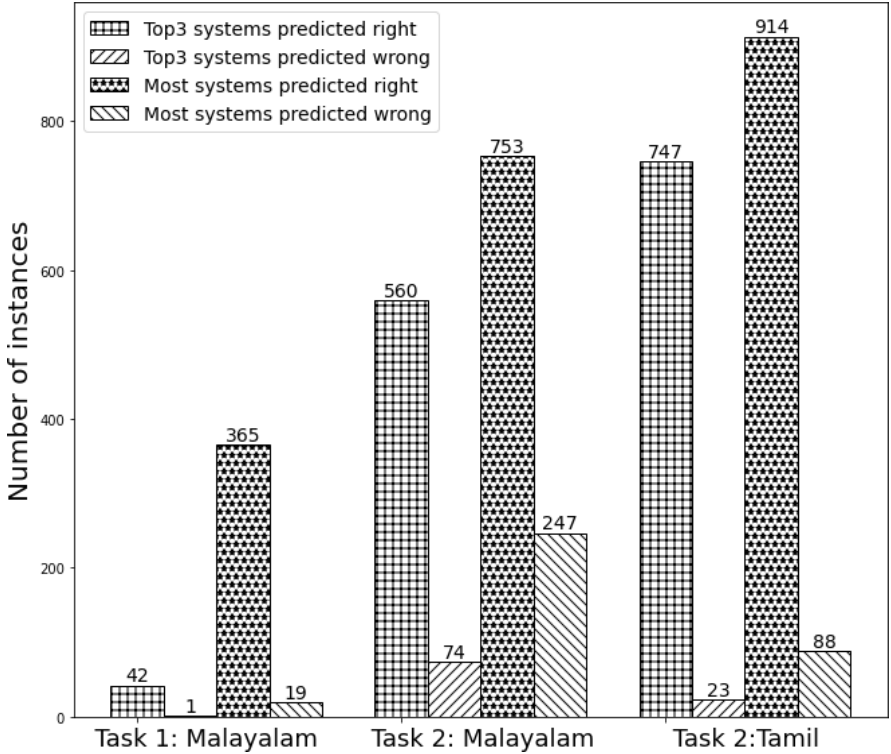

**Figure 1.** Performance analysis for HASOC subtasks

Regarding the HASOC task 2, there are two datasets including Malayalam and Tamil. Refer Table 9 for the distribution of data in the test set. In case of the Malayalam HASOC 2 dataset, there are 560 instances which are always predicted correctly by the top 3 systems. There 74 instances that are always predicted wrongly by the top 3 systems. There are 753 instances where all top 3 systems predicted correctly, except for one of the top 3 systems. Similarly there are 247 instances where where all top 3 systems predicted wrongly, except for one of the top 3 system. Recall that there are 1000 instances in the test set of HASOC 1 Malayalam dataset as in Table 9. In the HASOC task 2 Tamil dataset, there are are 747 instances which are always predicted correctly by the top 3 systems. There 23 instances that are always predicted wrongly by the top 3 systems. There are 914 instances where all top 3 systems predicted correctly, except for one of the top 3 systems. Similarly there are 88 instances where where all top 3 systems predicted wrongly, except for one of the top 3 system. Recall that there are 940 instances in the test set of HASOC 1 Tamil dataset as in Table 9.

The Dravidan langtech offensive language identification task includes three subtasks including Malayalam, Tamil, and Kannada. Figure 6.1 shows the results of analysis over the top 3 systems per subtask per language in the test set.

Regarding the Dravidian langtech Malayalam dataset, there are 1,540 instances which are always predicted correctly by the top 3 systems. There 460 instances that are always predicted wrongly by the top 3 systems. There are 1080 instances where all top 3 systems predicted correctly, except for one of the top 3 systems. Similarly there are 982 instances where where all top 3 systems predicted wrongly, except for one of the top 3 systems. Recall that there are 2001 instances in the Dravidian langtech Malayalam dataset as in Table 10.

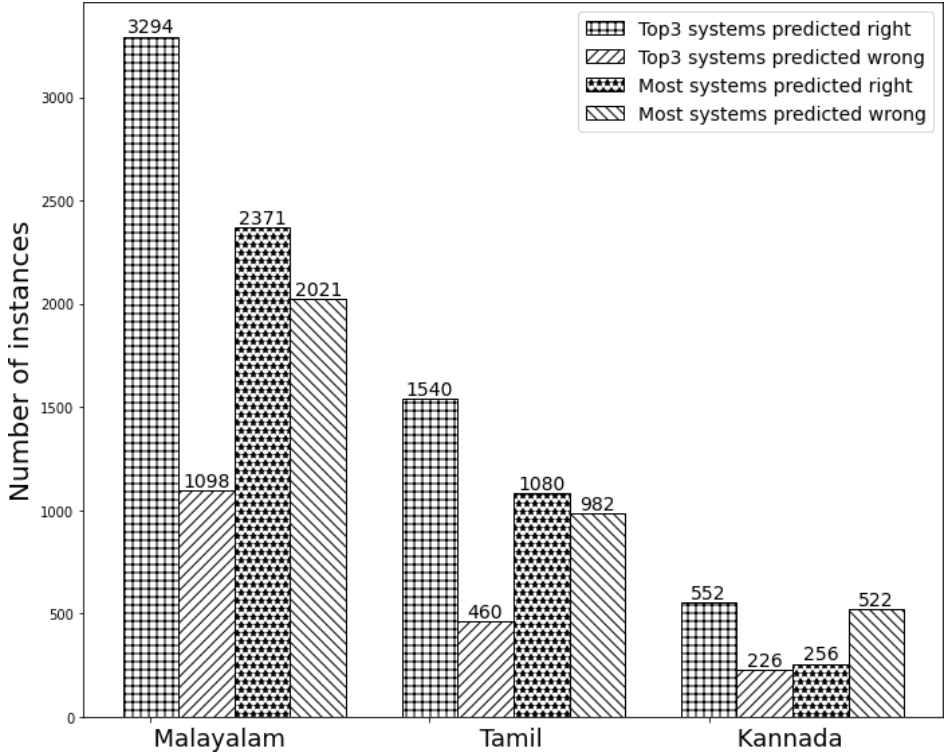

**Figure 2.** Performance analysis for Dravidian langtech

In case of the Dravidian langtech Tamil dataset, 3,294 instances are always correctly predicted by the top 3 systems. There are 1098 instances that is always predicted wrongly by the top 3 systems. There are 2371 instances where all top 3 systems predicted correctly, except for one of the top 3 systems. Similarly there are 2021 instances where where all top 3 systems predicted wrongly, except for one of the top 3 system. Recall that there are 4392 instances in the Dravidian langtech Malayalam dataset as in Table 10.

In the Dravidian langtech Kannada dataset, there are are 552 instances which are always predicted correctly by the top 3 systems. There 226 instances that are always predicted wrongly by the top 3 systems. There are 256 instances where all top 3 systems predicted correctly, except for one of the top 3 systems. Similarly there are 522 instances where where all top 3 systems predicted wrongly, except for one of the top 3 system. There are 778 instances in the test set as described in Table 10.

### *6.2 Qualitative error analysis*

In this section we present insights into the errors observed in the best performing models. We do not present a task specific error analysis as both the HASOC and Dravidian langtech shared tasks had the similar goal of identifying hate speech, despite having different labels.

**Microagression.** Microaggressions are subtle, often veiled, manifestations of human biases Breitfeller et al. (2019). The linguistic subtlety of microaggressions in communication has made it difficult for researchers to analyze their exact nature, and to quantify and extract microaggressions automatically. Microaggressions are offensive languages that even the state-of-the-art systems fail to identify.

ஆணவ ெகாைலகள் நடக்காவிடில் ...நம் ஐடயாளம் அளிக்கப்படும்... இயக்குனருக்கு வாழ்த்துக்கள்...தமிழ்நாடு முத்தைரயர்கள்

Gold label: Offensive

Prediction: Not-offensive

English translation: *If not for honor killing, our identity would be lost. Congratulations to the director.*

In the example above, the author of the comment subtly encourages violence. The context of the comment is in support of one's caste over another. Though they do not explicitly ask for killing someone from a different caste, they clearly expressing hate. Even the best performing systems suffer in these cases.

**World knowledge.** Systems fail to predict correctly as they do not possess the required world knowledge. In the following example,

*Sollitu poga vendiyathaana, 2 days Ku deactivate ah, varattum epdi kaluvi uuthurennumaatum paarunga*

Gold label: Offensive targeted

Prediction: Not-offensive

English translation: *Have they deactivated for two days? They should have told me before leaving. Wait and watch when I give them an earful.*

Thought the translation clearly shows hate towards someone the term *kaluvi uuthuren* is a colloquial metaphor used for scolding someone when it literally translates to *washing and pouring*. The classifiers fail to understand this colloquial metaphor due to the lack of world knowledge which explains their failure to predict correctly.

**Length of the sentences.** We observed that the length of the sentences played a vital role in prediction. That is classifiers failed when the length of the sentence is too short or too long. We present two examples one for each case.

*Puli murukan. Ithokke enth*

Gold label: Offensive

Prediction: Not offensive

English translation: *Puli Murugan (A movie title). What is this?*

In this example, the length of the sentence is too short and the classifiers fail to predict correctly.

*thala mattum than original mass...bcoz others papaerla vilambaram kotuthutu poraku trailar release panittu sathanai pannittatha pithikuvanga thala mttum than vilambarame seiyamal kalakkuvaar....thalada.. ..*

Gold label: Offensive untargeted

Prediction: Not offensive

English translation: *Thala (nickname of Actor Ajith Kumar) is the real hero. Other actors publish advertisements in Newspaper and then release trailer for the movie. Then they show off how successful they are. Thala never shows off but succeeds.*

Here, the length of the sentence is too long which makes the classifiers struggle to identify the category of the sentence.

**Negative words indicating positive sentiment.**

*mmale naattil degradarsinu oru kaalathhum kammi undaakilla...athaan nammala naadinte main shaapam*

Gold label: Not Offensive

Prediction: Offensive

English translation: *The biggest curse of our country is the people who degrade others.*

In this example, the author is frustrated by the hate speech others are uttering. However, the choice of words in his sentence such as shaapam (curse), confuses the classifier enabling them to make an incorrect prediction.

**Lack of enough data** We noticed some obvious instances where the sentence included hate words classifiers predicted an instance to be *Not offensive*. In these cases the instance clearly is *Offensive*. This misclassification has happened, despite the occurrence of many offensive words in the test instance. Specifically, this error type is observed predominantly in Kannada.

*@Vishal Shetty neenu beedhi soole maga antha gotthu bidu.... hogi beedhili nintko....*
Gold label: Not Offensive
Prediction: Offensive
English translation: *Go and Enjoy son of slut*

**Sarcasm Not detected** It was observed that the models could not capture sarcasm. Consider the Kannada example below.

*Śēkada 100 percent rasṭu viruses na China utpanna mādutte*
Gold label: Offensive Targeted Insult Group
Prediction: Not Offensive
English translation: *If it is 100 percent virus, then it is made in China*

The models fail to identify sarcasm, which is a complex linguistic phenomena. As sarcasm is one of the common way to express hate, the models might work better if trained with more sarcastic instances.

## 7. Conclusion and Future direction

This paper analyses the systems submitted for the HASOC shared tasks and DravidianLangTech workshop conducted in 2020. The shared task aims to develop machine learning models for identifying Malayalam, Tamil, and Kannada offensive posts on social media. The shared task has fostered corpora creation of code-mixed offensive content in Malayalam, Tamil, and Kannada. In case of the HASOC task we worked with Malyalam and Tamil datasets. In the case of the shared task in the DravidianLangTech workshop we worked with Tamil, Malayalam, and Kannada datasets.

The shared task motivates the researchers to develop machine learning models to identify Malayalam, Tamil, and Kannada offensive posts on social media and compare the results with other experts in this field. We conducted the HASOC task for 2021, and have observed that the number of participants doubled. This increase in the number of participants indicates the necessity of the shared task for Dravidian languages and the success of our previous shared tasks.

In the future, we intend to increase the size of the dataset. This increase in size would help the models to learn better on the instances were we observed the common errors discussed. For instance, we might be able to capture more hate speech that are non-keyword-based and are related to sarcasm. The detailed class-wise evaluation scores and the error analysis are also in our interest along with multilingual dataset. As systems trained on monolingual data fail on code-mixed data, strengthening the corpus for offensive language identification of code-mixed text in Dravidian languages is also inline. Further narrowing down offensive classes in the dataset with classes such as offensive which is not targeting anyone, offensive targeting individual, offensive targeting group, offensive targeting others are also in scope.

## References

**Ajees, A. P.** 2020. Ajees@HASOC-Dravidian-CodeMix-FIRE2020. In *FIRE (Working Notes)*.

**Andrew, J.** 2021. JudithJeyafreedaAndrew@DravidianLangTech-EACL2021:Offensive language detection for Dravidian Code-mixed YouTube comments. In *Proceedings of the First Workshop on Speech and Language Technologies for Dravidian Languages*. Association for Computational Linguistics.

**Arora, G.** 2020. Gauravarora@HASOC-Dravidian-CodeMix- FIRE2020: Pre-training ULMFiT on Synthetically Generated Code-Mixed Data for Hate Speech Detection. In *FIRE (Working Notes)*.

**Awatramani, V.** 2021. No Offense@DravidianLangTech-EACL2021: Offensive Tamil Identification and beyond the performance . In *Proceedings of the First Workshop on Speech and Language Technologies for Dravidian Languages*. Association for Computational Linguistics.

**Bali, K.**, **Sharma, J.**, **Choudhury, M.**, **and Vyas, Y.** 2014. "I am borrowing ya mixing ?" an analysis of English-Hindi code mixing in Facebook. In *Proceedings of the First Workshop on Computational Approaches to Code Switching*, pp. 116–126, Doha, Qatar. Association for Computational Linguistics.

**Balouchzahi, F.**, **B K, A.**, **and Shashirekha, H. L.** 2021. MUCS@DravidianLangTech-EACL2021:COOLI-Code-Mixing Offensive Language Identification. In *Proceedings of the First Workshop on Speech and Language Technologies for Dravidian Languages*. Association for Computational Linguistics.

**Barman, U.**, **Das, A.**, **Wagner, J.**, **and Foster, J.** 2014. Code mixing: A challenge for language identification in the language of social media. In *Proceedings of the First Workshop on Computational Approaches to Code Switching*, pp. 13–23, Doha, Qatar. Association for Computational Linguistics.

**Baruah, A.**, **Das, K. A.**, **Barbhuiya, F. A.**, **and Dey, K.** 2020. IIITG-ADBU@HASOC-Dravidian-CodeMix-FIRE2020: Offensive Content Detection in Code-Mixed Dravidian Text. In *FIRE (Working Notes)*.

**Benikova, D.**, **Wojatzki, M.**, **and Zesch, T.** 2018. What Does This Imply? Examining the Impact of Implicitness on the Perception of Hate Speech. In **Rehm, G. and Declerck, T.**, editors, *Language Technologies for the Challenges of the Digital Age*, pp. 171–179, Cham. Springer International Publishing.

**Bharathi, B. and Silvia A, A.** 2021. SSNCSE NLP@DravidianLangTech-EACL2021: Offensive Language Identification on Multilingual Code Mixing Text. In *Proceedings of the First Workshop on Speech and Language Technologies for Dravidian Languages*. Association for Computational Linguistics.

**Breitfeller, L.**, **Ahn, E.**, **Jurgens, D.**, **and Tsvetkov, Y.** 2019. Finding microaggressions in the wild: A case for locating elusive phenomena in social media posts. In *Proceedings of the 2019 Conference on Empirical Methods in Natural Language Processing and the 9th International Joint Conference on Natural Language Processing (EMNLP-IJCNLP)*, pp. 1664–1674, Hong Kong, China. Association for Computational Linguistics.

**Caldwell, R.** 1856. A comparative grammar of the Dravidian or south-Indian family of languages.(Madras: University of Madras. 1961).

**Chakravarthi, B. R.**, **Arcan, M.**, **and McCrae, J. P.** 2019. Comparison of Different Orthographies for Machine Translation of Under-Resourced Dravidian Languages. In *2nd Conference on Language, Data and Knowledge (LDK 2019)*, volume 70 of *OpenAccess Series in Informatics (OASIcs)*, pp. 6:1–6:14, Dagstuhl, Germany. Schloss Dagstuhl–Leibniz-Zentrum fuer Informatik.

**Chakravarthi, B. R.**, **Jose, N.**, **Suryawanshi, S.**, **Sherly, E.**, **and McCrae, J. P.** 2020a. A sentiment analysis dataset for code-mixed Malayalam-English. In *Proceedings of the 1st Joint Workshop on Spoken Language Technologies for Under-resourced languages (SLTU) and Collaboration and Computing for Under-Resourced Languages (CCURL)*, pp. 177–184, Marseille, France. European Language Resources association.

**Chakravarthi, B. R.**, **M, A. K.**, **McCrae, J. P.**, **B, P.**, **KP, S.**, **and Mandl, T.** 2020b. Overview of the track on hasoc-offensive language identification-dravidiancodemix. In *FIRE (Working Notes)*, pp. 112–120.

**Chakravarthi, B. R.**, **Muralidaran, V.**, **Priyadharshini, R.**, **and McCrae, J. P.** 2020c. Corpus creation for sentiment analysis in code-mixed Tamil-English text. In *Proceedings of the 1st Joint Workshop on Spoken Language Technologies for Under-resourced languages (SLTU) and Collaboration and Computing for Under-Resourced Languages (CCURL)*, pp. 202–210, Marseille, France. European Language Resources association.

**Chakravarthi, B. R.**, **Priyadharshini, R.**, **Jose, N.**, **Kumar M, A.**, **Mandl, T.**, **Kumaresan, P. K.**, **Ponnusamy, R.**, **R L, H.**, **McCrae, J. P.**, **and Sherly, E.** 2021. Findings of the shared task on offensive language identification in Tamil, Malayalam, and Kannada. In *Proceedings of the First Workshop on Speech and Language Technologies for Dravidian Languages*, pp. 133–145, Kyiv. Association for Computational Linguistics.

**Chakravarthi, B. R.**, **Rajasekaran, N.**, **Arcan, M.**, **McGuinness, K.**, **E.O'Connor, N.**, **and McCrae, J. P.** 2020d. Bilingual lexicon induction across orthographically-distinct under-resourced Dravidian languages. In *Proceedings of the Seventh Workshop on NLP for Similar Languages, Varieties and Dialects*, Barcelona, Spain.

**Chen, S. and Kong, B.** 2021. cs@DravidianLangTech-EACL2021: Offensive Language Identification Based On Multilingual BERT Model. In *Proceedings of the First Workshop on Speech and Language Technologies for Dravidian Languages*. Association for Computational Linguistics.

**Chittaranjan, G.**, **Vyas, Y.**, **Bali, K.**, **and Choudhury, M.** 2014. Word-level language identification using CRF: Code-switching shared task report of MSR India system. In *Proceedings of the First Workshop on Computational Approaches to Code Switching*, pp. 73–79, Doha, Qatar. Association for Computational Linguistics.

**Dave, B.**, **Bhat, S.**, **and Majumder, P.** 2021. IRNLPDAIICT@DravidianLangTech-EACL2021:Offensive Language identification in Dravidian Languages using TF-IDF Char N-grams and MuRIL . In *Proceedings of the First Workshop on Speech and Language Technologies for Dravidian Languages*. Association for Computational Linguistics.

**Dong, K.** 2020. YUN@HASOC-Dravidian-CodeMix-FIRE2020: A Multi-component Sentiment Analysis Model for Offensive Language Identification. In *FIRE (Working Notes)*.

**Dowlagar, S. and Mamidi, R.** 2021. OFFLangOne@DravidianLangTech-EACL2021: Transformers with the Class Balanced

**Edosomwan, S.**, **Prakasan, S. K.**, **Kouame, D.**, **Watson, J.**, **and Seymour, T.** 2011. The history of social media and its impact on business. *Journal of Applied Management and entrepreneurship*, 16(3):79–91.

**Garain, A.**, **Mandal, A.**, **and Naskar, S. K.** 2021. JUNLP@DravidianLangTech-EACL2021: Offensive Language Identification in Dravidian Langauges . In *Proceedings of the First Workshop on Speech and Language Technologies for Dravidian Languages*. Association for Computational Linguistics.

**Ghanghor, N.**, **Chakravarthi, B. R.**, **Priyadharshini, R.**, **Thavareesan, S.**, **and Krishnamurthy, P.** 2021. IIITK@DravidianLangTech-EACL2021: Offensive Language Identification and Meme Classification in Tamil, Malayalam and Kannada . In *Proceedings of the First Workshop on Speech and Language Technologies for Dravidian Languages*. Association for Computational Linguistics.

**Huang, B. and Bai, Y.** 2021. HUB@DravidianLangTech-EACL2021: Identify and Classify Offensive Text in Multilingual Code Mixing in Social Media. In *Proceedings of the First Workshop on Speech and Language Technologies for Dravidian Languages*. Association for Computational Linguistics.

**Jayanthi, S. M. and Gupta, A.** 2021. SJAJ@DravidianLangTech-EACL2021: Task-Adaptive Pre-Training of Multilingual BERT models for Offensive Language Identification. In *Proceedings of the First Workshop on Speech and Language Technologies for Dravidian Languages*. Association for Computational Linguistics.

**Jose, N.**, **Chakravarthi, B. R.**, **Suryawanshi, S.**, **Sherly, E.**, **and McCrae, J. P.** 2020. A Survey of Current Datasets for Code-Switching Research. In *2020 6th International Conference on Advanced Computing and Communication Systems (ICACCS)*.

**Kedia, K. and Nandy, A.** 2021. indicnlp@kgp@DravidianLangTech-EACL2021: Offensive Language Identification in Dravidian Languages . In *Proceedings of the First Workshop on Speech and Language Technologies for Dravidian Languages*. Association for Computational Linguistics.

**Keipi, T.**, **Näsi, M.**, **Oksanen, A.**, **and Räsänen, P.** 2016. *Online hate and harmful content: Cross-national perspectives*. Taylor & Francis.

**Krishnamurti, B.** 2003. *The Dravidian languages*. Cambridge University Press.

**Kumar, R.**, **Ojha, A. K.**, **Malmasi, S.**, **and Zampieri, M.** 2018. Benchmarking aggression identification in social media. In *Proceedings of the First Workshop on Trolling, Aggression and Cyberbullying (TRAC-2018)*, pp. 1–11, Santa Fe, New Mexico, USA. Association for Computational Linguistics.

**Kumar, Abhinav adn Saumya, S. and Singh, J. P.** 2020. NITP-AI-NLP@HASOC-Dravidian-CodeMix-FIRE2020: A Machine Learning Approach to Identify Offensive Languages from Dravidian Code-Mixed Text. In *FIRE (Working Notes)*.

**Li, Z.** 2021. Codewithzichao@DravidianLangTech-EACL2021: Exploring Multilingual Transformers for Offensive Language Identification on Code Mixing Text . In *Proceedings of the First Workshop on Speech and Language Technologies for Dravidian Languages*. Association for Computational Linguistics.

**Mandl, T.**, **Modha, S.**, **Kumar M, A.**, **and Chakravarthi, B. R.** 2020. Overview of the HASOC Track at FIRE 2020: Hate Speech and Offensive Language Identification in Tamil, Malayalam, Hindi, English and German. In *Forum for Information Retrieval Evaluation*, FIRE 2020, 29–32, New York, NY, USA. Association for Computing Machinery.

**McHugh, M. L.** 2012. Interrater reliability: the kappa statistic. *Biochemia medica*, 22(3):276–282. 23092060[pmid].

**Nair, S. and Fernandes, D.** 2021. professionals@DravidianLangTech-EACL2021. In *Proceedings of the First Workshop on Speech and Language Technologies for Dravidian Languages*. Association for Computational Linguistics.

**Pamungkas, E. W.**, **Basile, V.**, **and Patti, V.** 2020. Do you really want to hurt me? predicting abusive swearing in social media. In *Proceedings of the 12th Language Resources and Evaluation Conference*, pp. 6237–6246, Marseille, France. European Language Resources Association.

**Pathak, V.**, **Joshi, M.**, **Joshi, P.**, **Mundada, M.**, **and Joshi, T.** 2020. KBCNMUJAL@HASOC-Dravidian-CodeMix-FIRE2020: Using Machine Learning for Detection of Hate Speech and Offensive Codemix Social Media text. In *FIRE (Working Notes)*.

**Priyadharshini, R.**, **Chakravarthi, B. R.**, **Vegupatti, M.**, **and McCrae, J. P.** 2020. Named entity recognition for code-mixed Indian corpus using meta embedding. In *2020 6th International Conference on Advanced Computing and Communication Systems (ICACCS)*.

**Que, Q.**, **Wang, G.**, **and Jia, S.** 2021. Simon @ DravidianLangTech-EACL2021: Detecting Offensive Content in Kannada Language . In *Proceedings of the First Workshop on Speech and Language Technologies for Dravidian Languages*. Association for Computational Linguistics.

**Ranasinghe, T. and Zampieri, M.** 2020. WLV-RIT @ HASOC 2020: Offensive Language Identification in Code-switched Texts. In *FIRE (Working Notes)*.

**Renjit, S.** 2020. CUSAT-NLP@HASOC-Dravidian-CodeMix-FIRE2020: Identifying Offensive Language from Manglish Tweets. In *FIRE (Working Notes)*.

**Renjit, S. and Idicula, S. M.** 2021. CUSATNLP@DravidianLangTech-EACL2021:Language Agnostic Classification of Offensive Content in Tweets. In *Proceedings of the First Workshop on Speech and Language Technologies for Dravidian Languages*. Association for Computational Linguistics.

**Rudra, K.**, **Rijhwani, S.**, **Begum, R.**, **Bali, K.**, **Choudhury, M.**, and **Ganguly, N.** 2016. Understanding language preference for expression of opinion and sentiment: What do Hindi-English speakers do on Twitter? In *Proceedings of the 2016 Conference on Empirical Methods in Natural Language Processing*, pp. 1131–1141, Austin, Texas. Association for Computational Linguistics.

**Saha, D.**, **Paharia, N.**, **Chakraborty, D.**, **Saha, P.**, and **Mukherjee, A.** 2021. Hate-Alert@DravidianLangTech-EACL2021: Ensembling strategies for Transformer-based Offensive language Detection . In *Proceedings of the First Workshop on Speech and Language Technologies for Dravidian Languages*. Association for Computational Linguistics.

**Sai, S. and Sharma, Y.** 2020. Siva@HASOC-Dravidian-FIRE-2020: Multilingual Offensive Speech Detection in Code-mixed and Romanized Text. In *FIRE (Working Notes)*.

**Sakuntharaj, R. and Mahesan, S.** 2016. A novel hybrid approach to detect and correct spelling in Tamil text. In *2016 IEEE International Conference on Information and Automation for Sustainability (ICIAfS)*, pp. 1–6. IEEE.

**Sakuntharaj, R. and Mahesan, S.** 2017. Use of a novel hash-table for speeding-up suggestions for misspelt Tamil words. In *2017 IEEE International Conference on Industrial and Information Systems (ICIIS)*, pp. 1–5. IEEE.

**Sharif, O.**, **Hossain, E.**, and **Hoque, M. M.** 2021. NLP-CUET@DravidianLangTech-EACL2021: Offensive Language Detection from Multilingual Code-Mixed Text using Transformers. In *Proceedings of the First Workshop on Speech and Language Technologies for Dravidian Languages*. Association for Computational Linguistics.

**Singh, P. and Bhattacharyya, P.** 2020. CFILT IIT Bombay@HASOC-Dravidian-CodeMix FIRE 2020: Assisting ensemble of transformers with random transliteration. In *FIRE (Working Notes)*.

**Sivanantham, R. and Seran, M.** 2019. Keeladi: An Urban Settlement of Sangam Age on the Banks of River Vaigai. *India: Department of Archaeology, Government of Tamil Nadu, Chennai*.

**Sreelakshmi, K.**, **Premjith, B.**, and **Soman, K.** 2021. AmritaCENNLP@DravidianLangTech-EACL2021: Deep Learning-based Offensive Language Identification in Malayalam, Tamil and Kannada . In *Proceedings of the First Workshop on Speech and Language Technologies for Dravidian Languages*. Association for Computational Linguistics.

**Thavareesan, S. and Mahesan, S.** 2020. Word embedding-based Part of Speech tagging in Tamil texts. In *2020 IEEE 15th International Conference on Industrial and Information Systems (ICIIS)*, pp. 478–482.

**Tula, D.**, **Potluri, P.**, **MS, S.**, **Doddapaneni, S.**, **Sahu, P.**, **Sukumaran, R.**, and **Patwa, P.** 2021. Bitions@DravidianLangTech-EACL2021: Ensemble of Multilingual Language Models with Pseudo Labeling for Offense Detection in Dravidian Languages. In *Proceedings of the First Workshop on Speech and Language Technologies for Dravidian Languages*. Association for Computational Linguistics.

**Vasantharajan, C. and Thayasivam, U.** 2021. Hypers@DravidianLangTech-EACL2021: Offensive language identification in Dravidian code-mixed YouTube Comments and Posts. In *Proceedings of the First Workshop on Speech and Language Technologies for Dravidian Languages*. Association for Computational Linguistics.

**Veena, P.**, **Ramanan, P.**, and **Devi G, R.** 2020. CENMates@HASOC-Dravidian-CodeMix-FIRE2020: Offensive Language Identification on Code-mixed Social Media Comments. In *FIRE (Working Notes)*.

**Vikram, T. N. and Urs, S. R.** 2007. *Development of Prototype Morphological Analyzer for he South Indian Language of Kannada*, pp. 109–116. Springer Berlin Heidelberg, Berlin, Heidelberg.

**Yang, M.** 2021. Maoqin @ DravidianLangTech-EACL2021: The Application of Transformer-Based Model . In *Proceedings of the First Workshop on Speech and Language Technologies for Dravidian Languages*. Association for Computational Linguistics.

**Yasaswini, K.**, **Puranik, K.**, **Hande, A.**, **Priyadharshini, R.**, **Thavareesan, S.**, and **Chakravarthi, B. R.** 2021. IIITT@DravidianLangTech-EACL2021: Transfer Learning for Offensive Language Detection in Dravidian Languages . In *Proceedings of the First Workshop on Speech and Language Technologies for Dravidian Languages*. Association for Computational Linguistics.

**Zampieri, M.**, **Malmasi, S.**, **Nakov, P.**, **Rosenthal, S.**, **Farra, N.**, and **Kumar, R.** 2019. Predicting the type and target of offensive posts in social media. In *Proceedings of the 2019 Conference of the North American Chapter of the Association for Computational Linguistics: Human Language Technologies, Volume 1 (Long and Short Papers)*, pp. 1415–1420, Minneapolis, Minnesota. Association for Computational Linguistics.

**Zampieri, M.**, **Nakov, P.**, **Rosenthal, S.**, **Atanasova, P.**, **Karadzhov, G.**, **Mubarak, H.**, **Derczynski, L.**, **Pitenis, Z.**, and **Çöltekin, c.** 2020. SemEval-2020 Task 12: Multilingual Offensive Language Identification in Social Media (OffensEval 2020). In *Proceedings of SemEval*.

**Zhao, Y.** 2021. ZYJ123@DravidianLangTech-EACL2021: Offensive Language Identification based on XLM-RoBERTa with DPCNN. In *Proceedings of the First Workshop on Speech and Language Technologies for Dravidian Languages*. Association for Computational Linguistics.

**Zhu, Y. and Zhou, X.** 2020. Zyy1510@HASOC-Dravidian-CodeMix-FIRE2020: An Ensemble Model for Offensive Language Identification. In *FIRE (Working Notes)*.